\documentclass[]{entalpic}

\usepackage{amsmath}
\usepackage{amssymb}
\usepackage{graphicx}
\usepackage{natbib}
\usepackage[most]{tcolorbox}

\usepackage{enumitem}
\setlist{topsep=5pt, partopsep=5pt, itemsep=5pt, parsep=5pt}

\usepackage[utf8]{inputenc} % allow utf-8 input
\usepackage[T1]{fontenc}    % use 8-bit T1 fonts
\usepackage{hyperref}       % hyperlinks
\usepackage{url}            % simple URL typesetting
\usepackage{booktabs}       % professional-quality tables
\usepackage{amsfonts}       % blackboard math symbols
\usepackage{nicefrac}       % compact symbols for 1/2, etc.
\usepackage{microtype}      % microtypography
\usepackage{xcolor}         % colors
\usepackage{todonotes}
\usepackage{multirow}
\usepackage{amsmath}
\usepackage{siunitx}
\usepackage{caption}
\usepackage{graphicx}

\makeatletter
\renewcommand{\title}[1]{\gdef\@title{#1}\newcommand{\titlelist}{%
  {\LARGE\bfseries\rmfamily\raggedright\linespread{1.1}\selectfont #1\par}%
  \vspace{0.4cm}%
}}
\makeatother

\title{LeMat-Traj: A Scalable and Unified Dataset of Materials Trajectories for Atomistic Modeling}

\author[1,*]{Ali Ramlaoui}
\author[1]{Martin Siron}
\author[1]{Inel Djafar}
\author[1]{Joseph Musielewicz}
\author[1]{Amandine Rossello}
\author[1]{\\Victor Schmidt}
\author[1,*]{Alexandre Duval}

\affiliation[1]{Entalpic, Paris, France}
\contribution[*]{Corresponding authors}

\abstract{The development of accurate machine learning interatomic potentials (MLIPs) is limited by the fragmented availability and inconsistent formatting of quantum mechanical trajectory datasets derived from Density Functional Theory (DFT). These datasets are expensive to generate yet difficult to combine due to variations in format, metadata, and accessibility. To address this, we introduce LeMat-Traj, a curated dataset comprising over 120 million atomic configurations aggregated from large-scale repositories, including the Materials Project, Alexandria, and OQMD. LeMat-Traj standardizes data representation, harmonizes results and filters for high-quality configurations across widely used DFT functionals (PBE, PBESol, SCAN, r2SCAN). It significantly lowers the barrier for training transferrable and accurate MLIPs.
LeMat-Traj spans both relaxed low-energy states and high-energy, high-force structures, complementing molecular dynamics and active learning datasets. By fine-tuning models pre-trained on high-force data with LeMat-Traj, we achieve a significant reduction in force prediction errors on relaxation tasks. We also present LeMaterial-Fetcher, a modular and extensible open-source library developed for this work, designed to provide a reproducible framework for the community to easily incorporate new data sources and ensure the continued evolution of large-scale materials datasets. LeMat-Traj and LeMaterial-Fetcher are publicly available at \url{https://huggingface.co/datasets/LeMaterial/LeMat-Traj} and \url{https://github.com/LeMaterial/lematerial-fetcher}.}

\begin{document}

\maketitle

\section{Introduction}
The discovery and design of novel materials are essential for technological advancement, offering solutions to pressing global challenges such as sustainable energy and climate change mitigation~\citep{pyzer2022accelerating}. However, traditional lab experiments and computational approaches, particularly those involving Density Functional Theory (DFT), are resource-intensive \citep{zitnick2020introduction}. Machine Learning Interatomic Potentials (MLIPs) have emerged as a promising alternative, offering DFT-level accuracy at a fraction of the computational cost. This acceleration is crucial for enabling large-scale molecular dynamics (MD) simulations over long timescales and rapid exploration of material properties \citep{Unke_2021, duval2023faenet}, potentially fast-tracking the development of materials for applications like carbon capture, improved batteries, or more efficient catalysts. 

Graph Neural Networks (GNNs) have emerged as the most effective class of models for learning interatomic potentials, due to their ability to naturally represent atomic systems and to incorporate physical symmetries such as rotational and permutational equivariance \citep{duval2023hitchhiker}. As modern GNN architectures like EquiformerV2 \citep{liao_equiformerv2_2024} exhibit scaling laws behaviors \citep{brehmer2024doesequivariancematterscale}, the need for even larger, more diverse, and consistently processed datasets becomes predominant.
Despite several large-scale initiatives generating vast amounts of DFT data \citep{jain2013commentary, schmidt2024improving}, these datasets often remain siloed, employ distinct data formats, and use varying DFT parameters (e.g., functionals, parameters, pseudopotentials). This fragmentation poses a challenge for researchers aiming to leverage the full spectrum of available data, as combining these sources requires considerable preprocessing and harmonization efforts. Consequently, many MLIPs are trained on scattered and non-homogeneous datasets, potentially restricting their generalizability and predictive power, while introducing chemical bias due to the way the datasets are separately being used to train these models~\citep{schmidt2024improving}. Moreover, large architectures—some now comprising over 30 million parameters—stand to benefit from access to even bigger and more diverse datasets, as further scaling require proportionally more data to avoid overfitting and fully realize their expressive power \citep{liao_equiformerv2_2024}.

To overcome these limitations, we introduce LeMat-Traj, a large-scale, aggregated dataset of materials trajectories. LeMat-Traj compiles data from three prominent sources: Materials Project~\citep{jain2013commentary,jain2020materials}, Alexandria~\citep{schmidt2024improving}, and OQMD~\citep{saal2013materials}. It harmonizes these datasets into a unified format, encompassing calculations performed with various DFT functionals (PBE, PBESol, SCAN, and r2SCAN). Furthermore, we introduce \texttt{LeMaterial-Fetcher}, an open-source Python library designed for the systematic and reproducible curation of materials science datasets.
% The tool automates the fetching, transformation, validation, and formatting of data from diverse sources, facilitating continuous integration of new calculations and extension to other datasets or materials science tasks such as catalysis modeling.

\begin{figure}[t]
    \centering
    \includegraphics[width=1\linewidth]{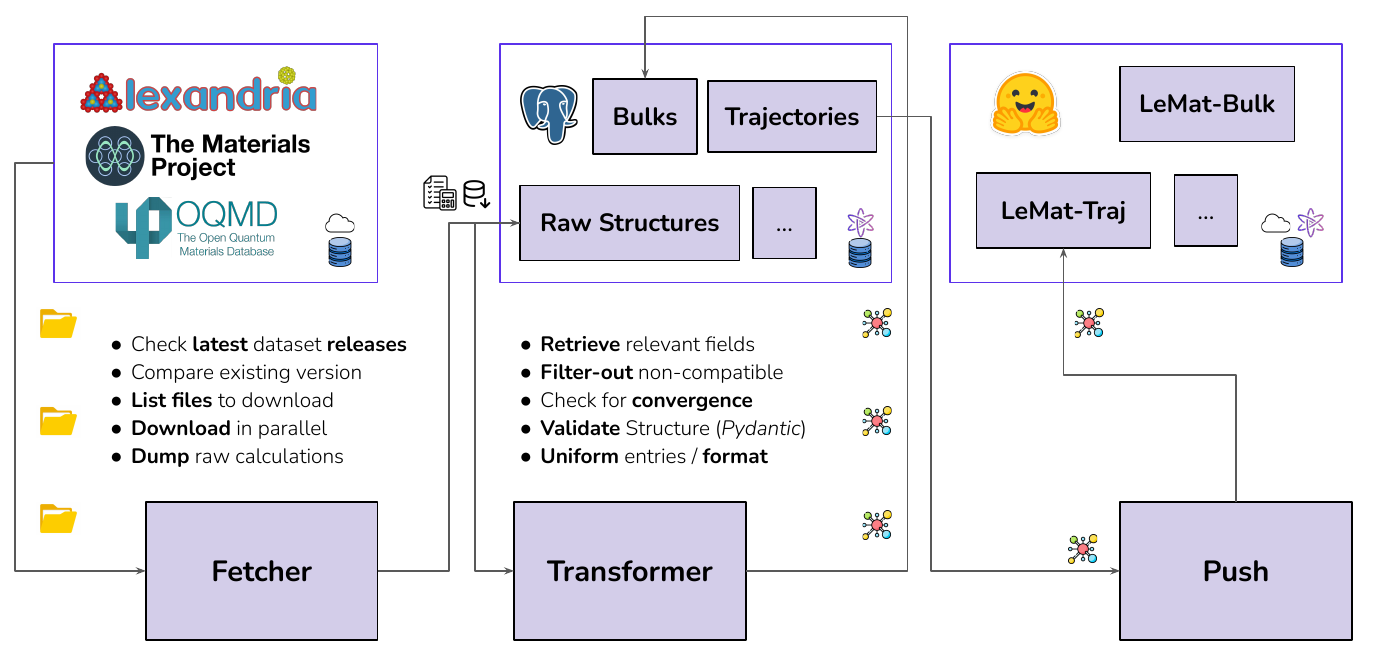}
    \caption{Data curation pipeline of LeMaterial-Fetcher. The library automates the process of fetching, transforming, validating, and harmonizing data from various sources, ensuring a consistent and reproducible dataset. The pipeline currently supports the continuous integration of fully relaxed bulk structures and full relaxation trajectories.}
    \label{fig:lemat_fetcher}
\end{figure}

Our contributions can be summarised as follows:
\begin{enumerate}
\item We release LeMat-Traj, to our knowledge one of the largest publicly available datasets of crystalline materials trajectories (120 million configurations). LeMat-Traj provides dense, high-quality coverage of near-equilibrium and low-force states—an underrepresented but crucial regime for accurate geometry optimization.
\item We empirically demonstrate the value of this data philosophy through extensive benchmarks. We show that by fine-tuning a MACE model with LeMat-Traj, we can reduce force prediction errors on relaxation tasks by over 36\% and improve performance on the Matbench Discovery stability benchmark by 10\%.
\item We introduce LeMaterial-Fetcher, a modular and extensible open-source library used to create LeMat-Traj. It provides a reproducible platform for community-driven curation, extension, and combination of large-scale materials datasets, enabling future research in multi-dataset and curriculum learning strategies. 
\end{enumerate}

We believe LeMat-Traj and LeMaterial-Fetcher will serve as a versatile foundation for the community, supporting not only the training of MLIPs but also a wide range of downstream tasks with crystalline materials, including benchmarking, subsampling strategies, self-supervised pretraining, and curriculum learning.

\section{Related Work}

The development of MLIPs has been closely correlated with the availability of suitable training datasets \citep{chanussot2020open, jain2020materials, levine2025open}. These datasets typically consist of sequences of atomic configurations, along with their corresponding energies and forces, generated from quantum mechanical simulations. Such sequences, often referred to as \textit{trajectories}, can originate from various simulation types, including geometry optimizations (tracing paths to energy minima) or molecular dynamics (MD, exploring configurations at specific thermodynamic conditions). Large-scale computational materials science initiatives like the Materials Project \citep{jain2013commentary, jain2020materials}, Alexandria \citep{schmidt2021alexandria, schmidt2024improving}, and the Open Quantum Materials Database (OQMD) \citep{saal2013materials}, along with resources like AFLOW \citep{eckert2024aflow}, NOMAD \citep{Draxl_2019}, and ColabFit \citep{vita2023colabfit}, have provided invaluable data to the community. 

While MLIPs are frequently trained using data derived from these sources such as MPtrj \citep{deng2023chgnet0} which curates relaxation trajectories from the Materials Project and has been used in models like CHGNet \citep{deng2023chgnet0}, MACE \citep{batatia2022mace0} and subsequent architectures, practitioners frequently encounter challenges~\citep{montes2022training}. Data from these diverse sources may employ different DFT parameters (e.g., functionals, k-points, pseudopotentials), varying data formats, and inconsistent preprocessing methodologies \citep{rossomme2023good}. This fragmentation means that combining data requires considerable, often repetitive, data engineering efforts \citep{wood2025family}, potentially limiting the generalizability and predictive power of the resulting MLIPs, and can introduce chemical biases depending on how individual datasets are leveraged~\citep{montes2022training}. 

% This challenge directly motivates the need for a systematically curated and harmonized dataset like LeMat-Traj, and an open-source tool like LeMaterial-Fetcher to build and maintain it along with other datasets like LeMat-Bulk \citep{siron2025lematbulk} for relaxed structures.

In parallel, two main philosophies of dataset design for training such MLIPs have emerged. One emphasizes broad exploration of the potential energy surface through high-force sampling, as in OMat24 \citep{barroso-luque2024open}, MatPES \citep{kaplan2025foundational} or MP-ALOE~\citep{kuner2025mp}, and active-learning datasets like ANI-1x \citep{smith2020ani}. These are well suited for pretraining robust models and capturing diverse regions of the configuration space~\citep{fu2022forces}. The other focuses on dense, near-equilibrium sampling from DFT geometry optimization trajectories, which provide clean, structured data in the low-force regime critical for accurate geometry optimization and stability prediction with datasets like Alexandria~\citep{schmidt2024improving}. Since machine learning force fields often display varying accuracy across the potential energy surface, with near-equilibrium and high-force regions posing different challenges \citep{vassilev2021challenges, loew2025universal}, these two philosophies of dataset design reflect complementary but compatible strategies to address that imbalance.

Recent work has underscored the need for large, harmonized, and extensible datasets that bridge these philosophies and mitigate fragmentation \citep{kaplan2025foundational, schmidt2024improving}. Our contribution follows this direction by introducing a systematically curated dataset of DFT trajectories together with an open-source pipeline to ensure reproducibility and extensibility. This places our work in line with ongoing efforts toward foundational datasets in materials science that can serve pretraining, benchmarking, and fine-tuning across a wide range of downstream MLIP applications.

Our work aims to address the data fragmentation challenge by providing not only a large, aggregated dataset but also a transparent, reproducible curation pipeline with LeMaterial-Fetcher. This aligns with the increasing need for foundational datasets in materials science \citep{kaplan2025foundational} that are large-scale, internally coherent, and extensible, facilitating pretraining, benchmarking, and fine-tuning across a wide range of downstream MLIP applications.

\section{Methodology}
LeMat-Traj is constructed by aggregating and processing data primarily from three major materials databases: Materials Project, Alexandria and OQMD (Open Quantum Materials Database). The core challenge lies in developing a scalable and reproducible methodology to handle the existing heterogeneity of these sources into a single and unified dataset.

\subsection{Unified Data Pipeline}
\label{sec:relaxation_number}
To address this, we developed LeMaterial-Fetcher, a highly parallelized Python-based open-source library described in Figure \ref{fig:lemat_fetcher}. It provides a unified and automated framework for:
\begin{itemize}
    \item \textbf{Fetching}: Interfacing with open APIs and direct downloads from various data sources.
    \item \textbf{Transformation}: Converting diverse input formats and attributes into a consistent schema. This includes standardizing atomic structure representations, energy units, and force components. It also handles the extraction and organization of metadata related to DFT calculations. All of this is done by allowing to interface with powerful atomistic modelling tools like \texttt{Pymatgen} \citep{ONG2013314}, \texttt{Matminer} \citep{WARD201860}.
    \item \textbf{Validation}: Implement checks to ensure data quality and integrity, such as verifying physical plausibility or consistency across reported values.
    \item \textbf{Harmonization}: Aligning DFT calculation parameters where possible and creating separate splits of data based on key parameters like the DFT functional.
    \item \textbf{Push}: Exporting the curated dataset in a user-friendly and efficient format, for direct use with libraries like HuggingFace's \texttt{Datasets} \citep{lhoest2021datasets}. This allows for easy integration with existing ML frameworks and tools, because they can adapt to limited computational resources, but also data versioning and metadata tracking as outlined in \citep{Draxl_2019}.
\end{itemize}
LeMaterial-Fetcher is designed to be modular, extensible but also scalable and fast, allowing for the easy integration of new data sources (e.g., future integration of quantum calculations sources) and adaptation to more materials science domains such catalysis, experiments, defects.
This framework ensures the reproducibility of LeMat-Traj and facilitates continuous integration of new DFT calculations as they become available from the source databases. It eliminates the need to manually iterate through datasets, download them, and then apply updates before releasing new versions. Additional details on the pipeline design are provided in Appendix~\ref{app:lemat_fetcher}.

\subsection{Data Sources and Harmonization}
LeMat-Traj specifically extracts geometry optimization trajectories from DFT calculations. A key aspect of our curation is the harmonization of data across different exchange–correlation functionals. We categorize trajectories based on the reported functional, primarily focusing on PBE, PBESol, SCAN, and r2SCAN, allowing users to train functional-specific models or to explore multi-fidelity learning across levels of theory (section~\ref{sec:benchmarks}). Table~\ref{tab:functional_data_summary} gives a full summary of the dataset partitioning.

The dataset follows the OPTIMADE specification \citep{andersen2021optimade}, enabling interoperability with other datasets that follow the same standard. We introduce a slight adaptation to accommodate trajectory data: each entry in the database corresponds to an individual atomistic configuration, which is part of a trajectory and is associated with energy and force information. Full optimization trajectories can be reconstructed by grouping entries by a shared trajectory identifier. This design choice facilitates seamless integration into machine learning interatomic potential (MLIP) training pipelines, where per-frame forces and energies are required.

To support trajectory-specific use cases, two new fields are introduced into the schema:
\begin{enumerate}
    \item \textbf{Relaxation Step}: An integer indicating the step number of the structure within a given geometry optimization sequence.
    \item \textbf{Relaxation Number}: An identifier that distinguishes different optimization runs for the same initial structure. This is particularly useful in high-throughput settings, where structures may undergo coarse relaxations before being re-relaxed with tighter thresholds or more accurate methods.
\end{enumerate}

\begin{table}[ht]
\centering
\caption{Number of trajectories and atomic configurations per source database and functional.}
\begin{tabular}{llcc}
\toprule
\textbf{Functional} & \textbf{Database} & \textbf{Number of Trajectories} & \textbf{Number of configurations} \\
\midrule
\multirow{3}{*}{PBE} 
& Materials Project & 195,721 & 3,649,785 \\
& Alexandria        & 3,414,074 & 110,804,226 \\
& OQMD              & 135,966 & 264,782 \\
\midrule
\multirow{2}{*}{PBESol} 
& Materials Project & 39,981 & 309,873 \\
& Alexandria        & 252,791 & 6,099,623 \\
\midrule
SCAN 
& Materials Project & 7,756 & 180,528 \\
\midrule
r2SCAN
& Materials Project & 37,888 & 516,576 \\
\bottomrule
\end{tabular}
\label{tab:functional_data_summary}
\end{table}

\subsection{Data Filtering}
\label{sec:data_filtering}
Our data filtering strategy prioritizes retaining a large volume of diverse configurations while establishing quality control. To this end, several criteria were applied:
First, any atomic configuration lacking either energy or atomic force data was discarded.
Second, entire trajectories were removed if the energy difference between the penultimate and final optimization step exceeded a threshold of $2 \times 10^{-2}$ eV, a criterion adapted from MPtrj \citep{deng2023chgnet0} to ensure reasonable convergence.
Third, trajectories were also excluded if the maximum atomic force norm in the final configuration surpassed $0.2 \, \text{eV/\AA}$, i.e. the structure is not fully relaxed. While this force threshold is relatively high, it allows the inclusion of structures that, despite not being fully relaxed, still provide valuable information about the potential energy surface far from equilibrium, enriching the dataset for training robust force fields.
Finally, all configurations were validated against the OPTIMADE format specifications, and any entry failing these schema checks or other implemented validation tests was removed.

\subsection{Alternative training tasks}
The trajectory data and associated metadata in LeMat-Traj support the exploration of training tasks beyond standard force and energy prediction.

\paragraph{Direct Structure-to-Property Prediction and Amortized Optimization.}
LeMat-Traj is suitable for Initial Structure to Relaxed Structure/Energy (IS2RE/IS2RS) tasks \citep{chanussot2020open}, as each trajectory contains the initial unrelaxed configuration, the final relaxed state, and its energy. This data structure can be used for developing \textit{amortized optimization} methods for crystal structure relaxation \citep{amos2022tutorial}. In contrast to MLIPs that provide forces for an external optimizer, amortized methods attempt to learn the direct mapping from an initial structure to its relaxed state by utilizing the DFT optimization paths within the dataset. Such approaches may be beneficial for applications requiring rapid structure prediction, for example, in high-throughput screening or for large systems where conventional relaxation methods can be computationally demanding \citep{ase-paper}. While not impossible with MPtrj and Alexandria, the raw format of these datasets makes this task difficult. In contrast, the relaxation step number associated to each trajectory, and the name of the trajectory it belongs can be easily leveraged for this specific task on LeMat-Traj.

\paragraph{Self-Supervised Learning (SSL) for Representation Learning.}
The scale and diversity of LeMat-Traj also make it a relevant dataset for pre-training models using self-supervised learning (SSL) techniques \citep{miret2025energy}. The sequential information in trajectories, the relationships between different configurations along a relaxation path, and the large number of atomic configurations can serve as signals for SSL. For example, methods based on contrastive learning (e.g. DeNS \citep{liao2024generalizing}), masked atom or coordinate prediction, or generative pre-training (such as diffusion models, e.g. ORB \citep{neumann2024orb0}) could be applied. Learning to predict masked information or reconstruct parts of the input structures can help models develop general atomic representations. These representations could then be used as a starting point for fine-tuning on specific downstream tasks, potentially aiding sample efficiency and generalization, analogous to approaches in other domains like natural language processing \citep{devlin2019bertpretrainingdeepbidirectional}. The consistent formatting of LeMat-Traj facilitates the application of these SSL methods.

The unified format produced by LeMaterial-Fetcher allows for the distribution of LeMat-Traj via platforms like HuggingFace Datasets, providing access to the data for these training approaches.

\section{Coverage of Chemical and Configurational Space}
LeMat-Traj comprises approximately 120 million atomic structures derived from geometry optimization trajectories. The dataset is partitioned based on the DFT functional used for the calculations: PBE, PBESol, SCAN, and r2SCAN. This partitioning facilitates targeted model training and research into multi-fidelity approaches.

\subsection{Chemical and Structural Diversity}
\begin{figure}[h]
  \centering
    \includegraphics[width=1\linewidth]{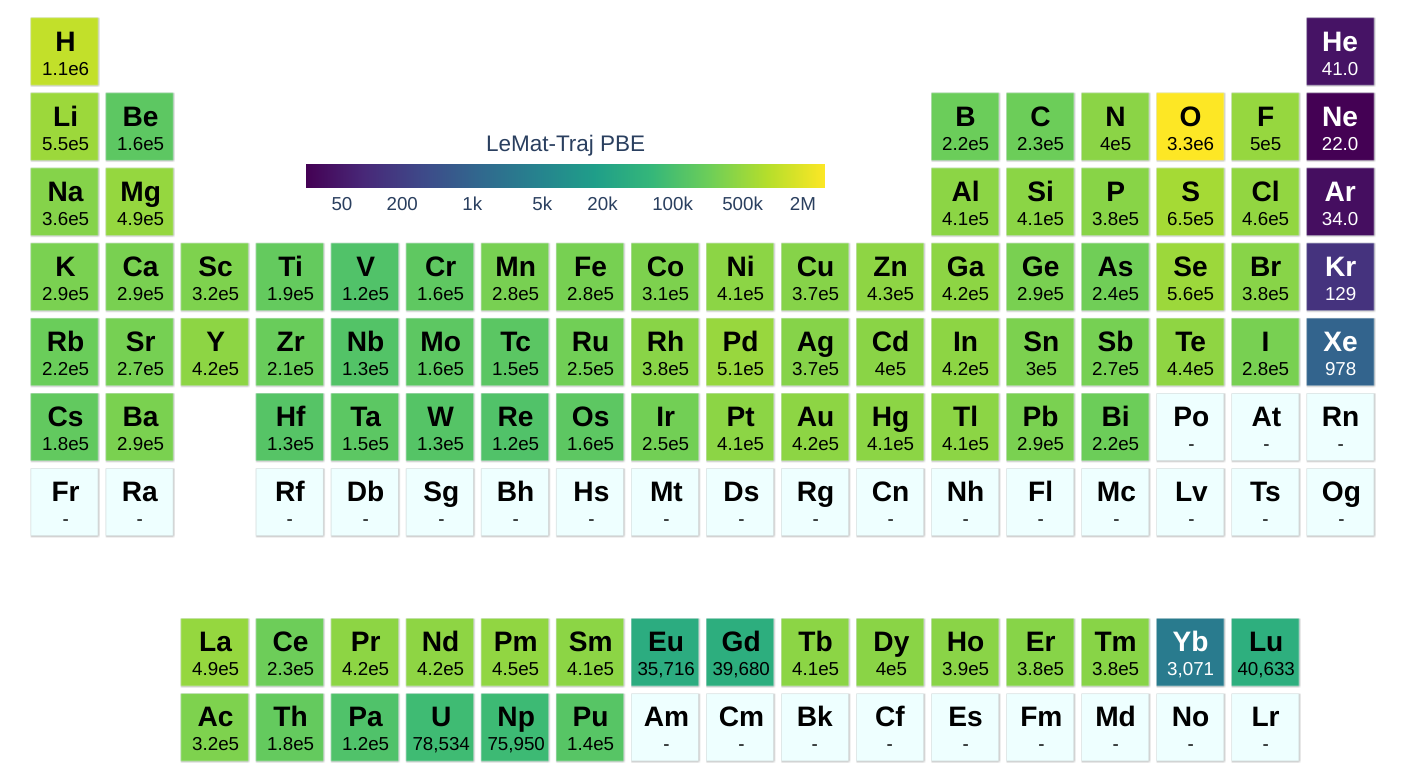}
  \caption{Chemical distribution in number of trajectories for the PBE split of LeMat-Traj using \texttt{Pymatviz}~\citep{riebesell_pymatviz_2022}.}
  \label{fig:chemical_formula_distribution}
\end{figure}

We compare the elemental and structural diversity of LeMat-Traj with other popular datasets such as MPtrj \citep{deng2023chgnet0} and MatPES \citep{kaplan2025foundational}.
LeMat-Traj aims to offer a broader coverage by combining multiple sources as illustrated in Figure~\ref{fig:chemical_formula_distribution}. While MPtrj primarily focuses on Materials Project data, LeMat-Traj's explicit harmonization and inclusion of OQMD and Alexandria data offer a unique combination of scale and more balanced distribution.

While the Alexandria dataset constitutes the majority of the PBE split by volume (approx. 92\%), the inclusion of data from Materials Project and OQMD is critical for diversity. First, it enriches the chemical space; Materials Project contains a higher concentration of oxides and battery materials, balancing the bi-metallic bias present in Alexandria. Second, it diversifies the force distribution; the average maximum force norm in Materials Project trajectories is significantly higher (593 meV/Å) than in the rest of the dataset (110 meV/Å), providing crucial high-force examples that help prevent models from under-estimating forces during relaxation.

\paragraph{Inclusion of Equilibrium Structures.}
A notable feature of LeMat-Traj is the inclusion of equilibrium structures from OQMD, which is rarely leveraged by ML practitioners when training machine-learned interatomic potentials (MLIPs). These configurations, characterized by near-zero atomic forces, serve as valuable reference points for MLIPs, particularly in capturing energy minima accurately. While relaxation trajectories naturally include low-force structures near convergence, the explicit addition of a large and diverse set of OQMD equilibrium configurations enhances the dataset’s richness. Although these single-point structures may be underrepresented compared to the total number of frames in full trajectories, they can be strategically leveraged by models focused on accurately learning stable configurations.

\subsection{Trajectory Analysis}

\begin{figure}[h]
    \centering
    \includegraphics[width=0.7\linewidth]{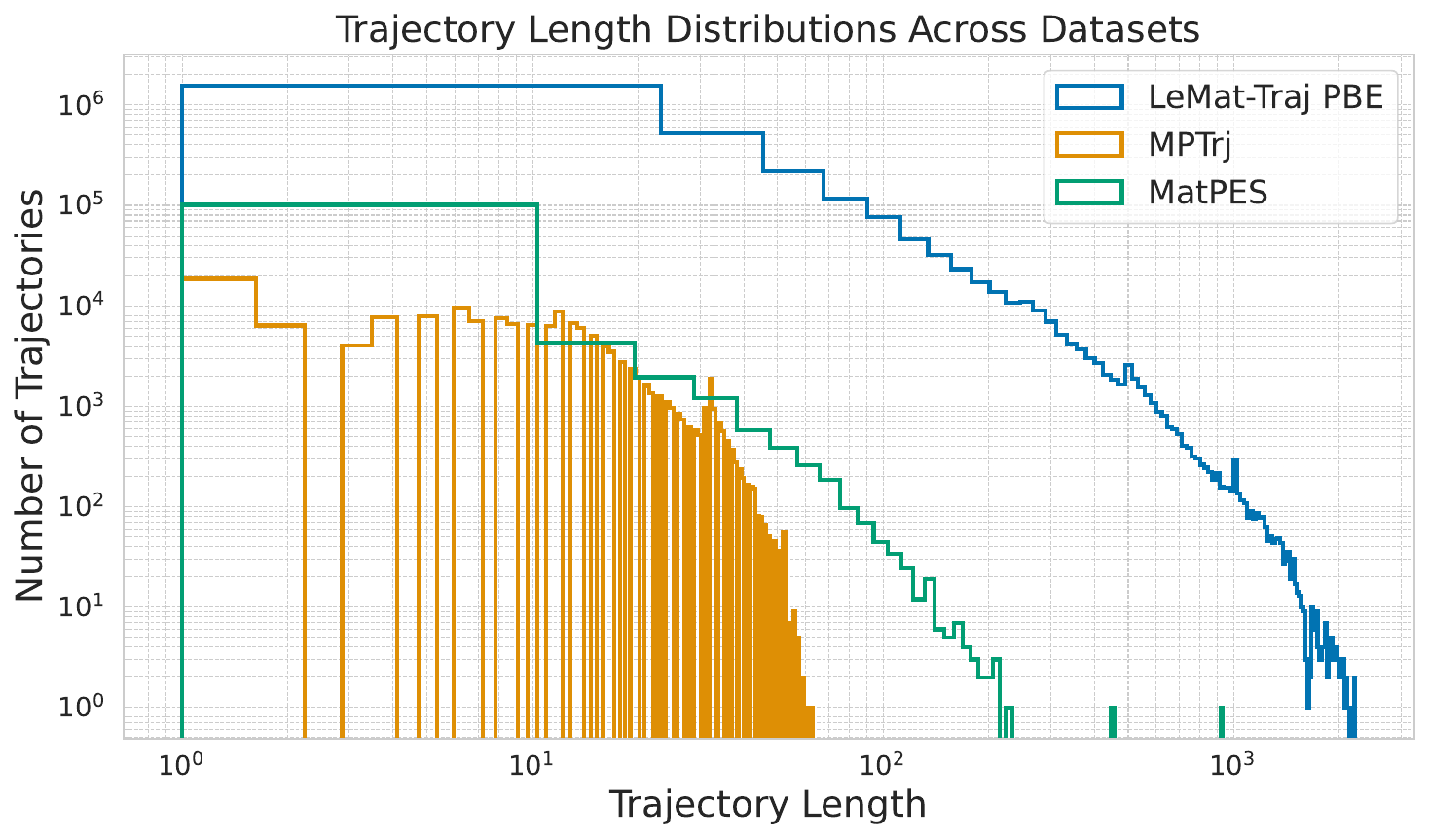}
    \caption{Comparison of trajectory length distributions for LeMat-Traj (PBE split), MPtrj, and MatPES, on a log-log scale. For every trajectory, the number of configurations associated is computed. LeMat-Traj exhibits a broader range of trajectory lengths.}
    \label{fig:trajectory_length}
\end{figure}

\paragraph{Trajectory Length.}
Figure~\ref{fig:trajectory_length} shows the distribution of trajectory lengths in LeMat-Traj. 
LeMat-Traj exhibits a broad distribution, with many trajectories across all length scales. It uniquely features a long tail with a significant number of trajectories extending beyond 100 frames, and even exceeding 1000 frames. In contrast, MPtrj is predominantly characterized by shorter trajectories, with the majority having fewer than 50 frames and a pronounced spikiness in its distribution at very short lengths. MatPES shows a broader distribution than MPtrj, with more medium-length trajectories (up to 100-200 frames), but still lacks the extensive representation of very long trajectories seen in LeMat-Traj. These longer trajectories are not indicative of optimization issues but are rather a feature of the highly stringent convergence criteria used in the source calculations, representing valid, but slow, convergence paths to energy minima.

\paragraph{Targets spread along trajectories.}

\begin{figure}[h]
  \centering
  \begin{subfigure}{0.47\linewidth}
    \includegraphics[width=\linewidth]{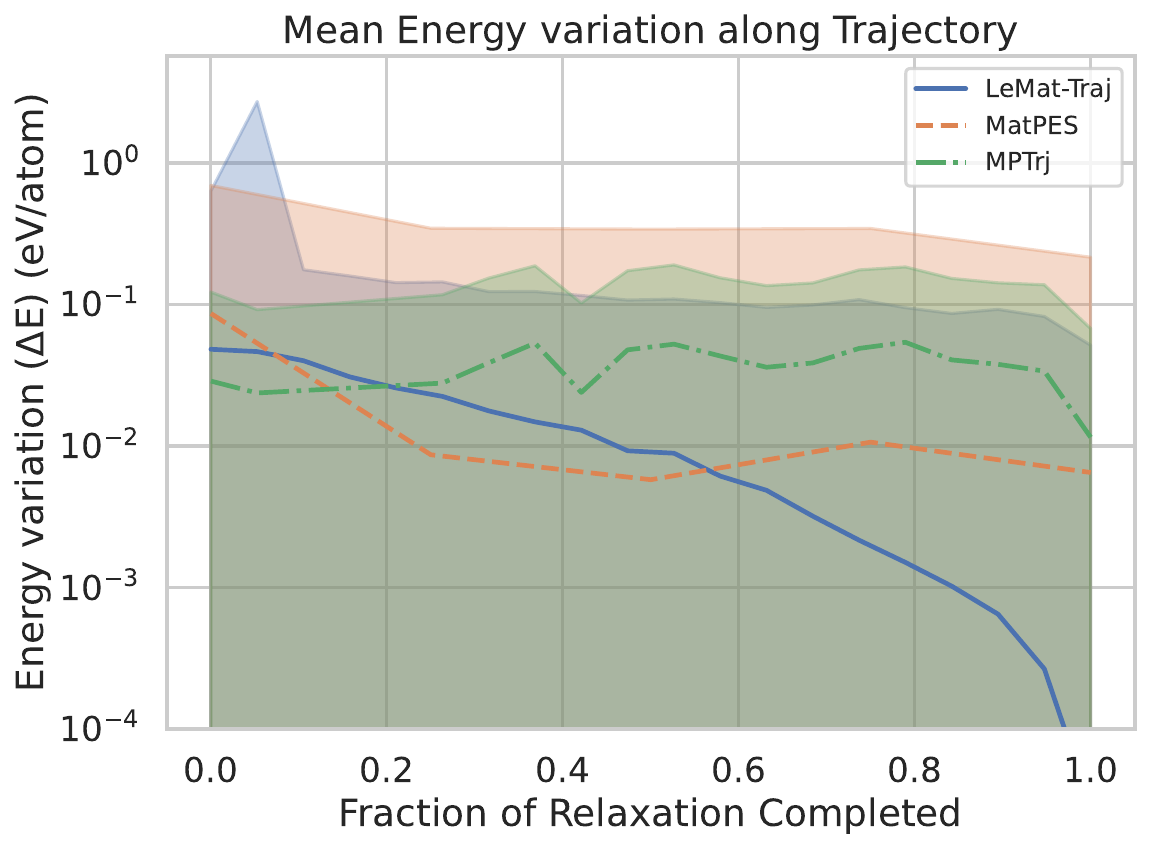}
    \caption{$\Delta E = E^t - E^T$}
    \label{fig:energy_variation}
  \end{subfigure}
  \hfill
  \begin{subfigure}{0.47\linewidth}
    \includegraphics[width=\linewidth]{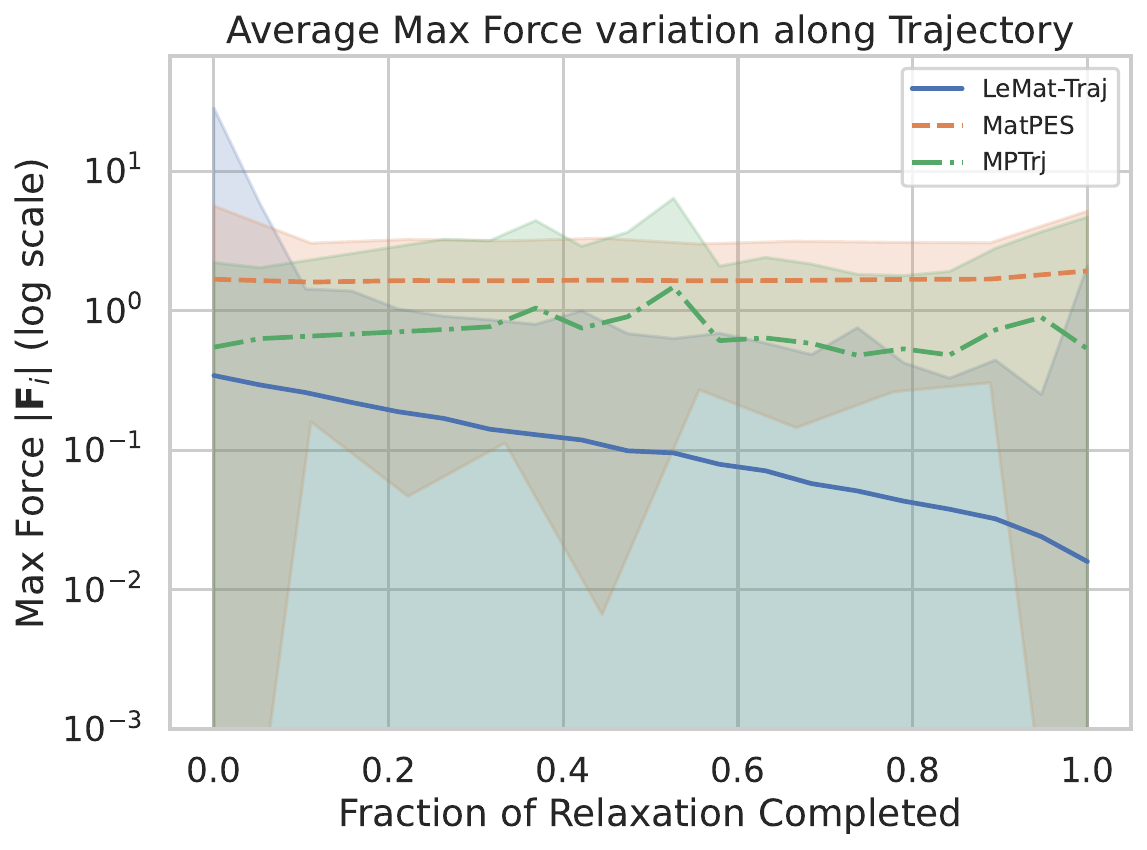}
    \caption{Maximum Force Norm}
    \label{fig:force_variation}
  \end{subfigure}
  \caption{Evolution of mean energy variation ($\Delta E = E^t - E^T$, where $E^t$ is current step energy and $E_T$ is final relaxed energy) per atom (a) and average maximum atomic force (b) as a function of the fraction of relaxation completed. Trajectories from LeMat-Traj, MatPES, and MPtrj are binned by their normalized progress. Solid lines represent the mean values, and shaded areas depict one standard deviation, both on a logarithmic y-axis. LeMat-Traj demonstrates comprehensive sampling from high-energy/high-force initial states to well-converged, low-energy/low-force final states.}
  \label{fig:trajectory_comp_datasets}
\end{figure}

Figure~\ref{fig:trajectory_comp_datasets} illustrates the evolution of mean energy variation ($\Delta E$ relative to the final relaxed state) and average maximum atomic forces norm throughout the relaxation trajectories of LeMat-Traj, MatPES, and MPtrj. LeMat-Traj uniquely demonstrates comprehensive sampling across the entire relaxation pathway. At the initial stages (low fraction of relaxation completed), it encompasses a wide distribution of high-energy and high-force configurations, with mean $\Delta E$ around 0.05 eV/atom (and variance extending >1 eV/atom from structures that are very far from their relaxed states iniially) and mean maximum forces around 0.3-0.4 eV/Å (variance extending >1 eV/Å). Crucially, as relaxations progress towards completion, LeMat-Traj systematically converges to very low $\Delta E$ (approaching $10^{-3}-10^{-4}$ eV/atom) and near-zero maximum forces (mean ~0.01-0.02 eV/Å, with significant density below $10^{-3}$ eV/Å). This shows a robust sampling both far-from-equilibrium states and accurately representing near-equilibrium energy minima and low-force structures, making LeMat-Traj well-suited for training versatile MLIPs capable of both high accuracy for stable configurations and robustness across diverse energy landscapes.

\subsection{Potential Energy Surface}

\begin{figure*}[h]
  \centering
  \begin{subfigure}{0.475\linewidth}
    \includegraphics[width=\linewidth]{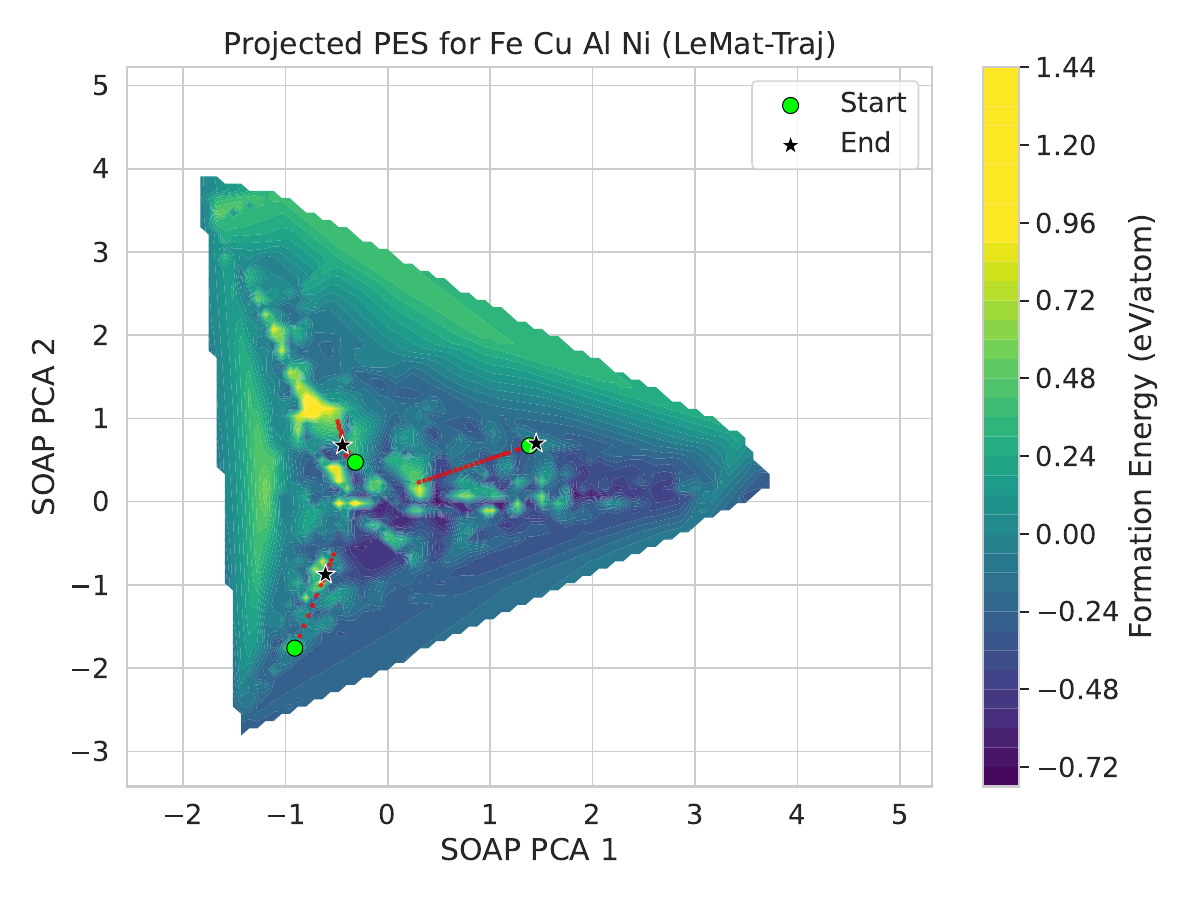}
    \caption{LeMat-Traj PBE}
    \label{fig:lemat_pes}
  \end{subfigure}
  \hfill
  \begin{subfigure}{0.475\linewidth}
    \includegraphics[width=\linewidth]{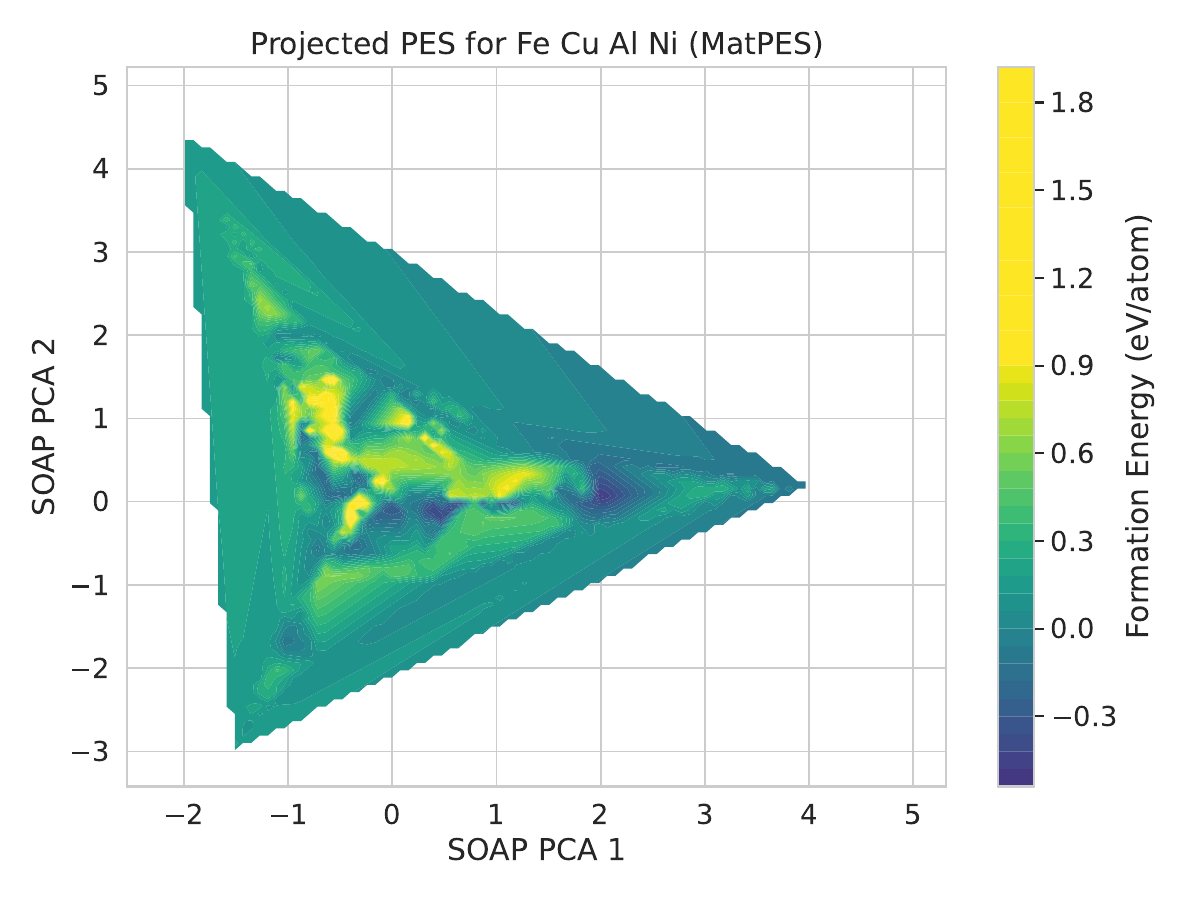}
    \caption{MatPES}
    \label{fig:matpes_pes}
  \end{subfigure}
  \caption{Projected Potential Energy Surfaces (PES) for the metallic Fe-Cu-Al-Ni systems.
Atomic configurations are featurized using SOAP descriptors~\citep{HIMANEN2020106949} and projected onto their first two principal components. The PCA 1 and PCA 2 axes are qualitative representations of structural similarity and do not have a direct physical interpretation.
Color indicates formation energy (eV/atom).
\textbf{(a)} PES derived from the LeMat-Traj PBE dataset. Green circles and black stars mark initial and final structures of example trajectories (red lines). The visualization highlights LeMat-Traj's dense, high-frequency sampling of the PES, which is crucial for resolving fine details near energy minima.
\textbf{(b)} PES derived from the MatPES dataset, showing a broader but sparser sampling of the overall landscape.
}
    \label{fig:pes_comparison}
\end{figure*}

To visualize the coverage of the potential energy surface (PES) by LeMat-Traj, we projected atomic configurations onto a lower-dimensional space derived from Smooth Overlap of Atomic Positions (SOAP) descriptors~\cite{HIMANEN2020106949}. Figure~\ref{fig:pes_comparison} illustrates this for the systems in the metallic Fe-Cu-Al-Ni hull within the PBE functional subset of LeMat-Traj, contrasting it with a similar projection for the MatPES dataset. LeMat-Traj projection (\ref{fig:lemat_pes}) reveals a broad exploration of the PES, with example trajectories (red lines) originating from diverse initial high-energy states (green circles) and converging towards distinct low-energy minima (black stars). The gradient energy gradient is clearly visible in the line levels far from the very high energy regions. This visualization is  also very similar with the MatPES projection (\ref{fig:matpes_pes}) which, while also covering a significant area, appears to have a different structural sampling emphasis, with less granularity around maxima, revealing a smaller number of saddle points. Further details on the visualization methodology are provided in Appendix~\ref{app:pes}.

\section{Results}
\label{sec:benchmarks}
To empirically validate the utility of LeMat-Traj, we conduct a series of benchmark experiments using the MACE architecture \citep{batatia2022mace0}, a well-established and performant equivariant model. These experiments are designed to demonstrate the dataset's value for improving model accuracy on relaxation-focused tasks, both through fine-tuning and in downstream applications.
A key hypothesis of our work is that LeMat-Traj's dense sampling of near-equilibrium states is complementary to datasets focused on high-force configurations. While high-force data, such as in OMat24, provides strong gradients that facilitate stable initial training and learning of the general energy landscape, LeMat-Traj is designed to refine model accuracy in the low-force regime critical for geometry optimization.

\subsection{Complementary Value for Fine-Tuning}
To test our hypothesis, we evaluate the performance of a MACE model pre-trained on the general-purpose OMat24 dataset and then fine-tune it on LeMat-Traj. As shown in Table~\ref{tab:finetuning_benchmark}, while the OMat24-trained model serves as a strong baseline, fine-tuning on LeMat-Traj reduces the Force MAE on our held-out test set of relaxation trajectories by over 36\%. This result provides direct evidence that LeMat-Traj contains critical information for achieving high fidelity in force predictions near energy minima, a crucial capability for accurate geometry optimization.
\begin{table}[h]
\centering
\caption{Performance of MACE on the LeMat-Traj PBE 10K held-out test set. Fine-tuning a model pre-trained on OMat24 with LeMat-Traj significantly reduces prediction errors, demonstrating the complementary nature of the datasets.}
\label{tab:finetuning_benchmark}
\begin{tabular}{lccc}
\toprule
\textbf{MACE Training Dataset} & \textbf{Energy MAE (meV) ↓} & \textbf{Force MAE (meV/Å) ↓} & \textbf{Force Cos ↑} \\
\midrule
OMat24 & 59.5 & 42.7 & 0.29 \\
LeMat-Traj only & 25.3 & 50.8 & 0.23 \\
\textbf{OMat24 + ft LeMat-Traj} & \textbf{18.8} & \textbf{27.2} & \textbf{0.30} \\
\bottomrule
\end{tabular}
\end{table}
\subsection{Downstream Performance on Matbench Discovery}
To assess practical utility, we evaluate our models on a subset of the Matbench Discovery benchmark, which measures a model's ability to predict the stability of novel crystalline materials. This task relies heavily on accurate structural relaxation. Table~\ref{tab:matbench_benchmark} shows that the MACE model trained on a split of LeMat-Traj with left-out matching protocol from Matbench Discovery following the method in~\citet{barroso-luque2024open} significantly outperforms the same model architecture trained on OMat24 or MPtrj alone, achieving a 10\% higher F1 score. The best performance is achieved by the model pre-trained on OMat24 and fine-tuned on LeMat-Traj, reinforcing the value of combining high-force and near-equilibrium data,

\begin{table}[h]
\centering
\caption{Matbench Discovery benchmark results on a 50k uniform subset. Models incorporating LeMat-Traj data achieve superior performance in predicting material stability.}
\label{tab:matbench_benchmark}
\begin{tabular}{lccc}
\toprule
\textbf{Model (Training Set)} & \textbf{F1 Score ↑} & \textbf{MAE (meV) ↓} & \textbf{RMSE (meV) ↓} \\
\midrule
MACE (OMat24) & 0.575 & 87.8 & 172.8 \\
MACE (MPtrj) & 0.694 & 47.2 & 83.9 \\
MACE (LeMat-Traj Full) & 0.768 & 37.2 & 69.0 \\
\textbf{MACE (OMat24 + ft-LeMat-Traj)} & \textbf{0.772} & \textbf{33.4} & \textbf{67.8} \\
\bottomrule
\end{tabular}
\end{table}

\subsection{Multi-Fidelity Learning.}
A notable challenge in materials modeling is the transferability of MLIPs trained on data from one level of theory (e.g., a specific DFT functional) to another. LeMat-Traj, with standardized formats of its different splits for PBE, PBESol, SCAN, and r2SCAN, provides a natural testbed for multi-fidelity learning strategies. 
We conduct experiments to assess how well models trained on one functional (e.g., PBE) can be fine-tuned or adapted for tasks involving another functional (e.g., PBESol and r2SCAN). 
For each of the PBESol and r2SCAN datasets, we use the subset described in Appendix~\ref{app:subset} during the experiments.
\begin{enumerate}
    \item We train a MACE model from scratch (using the same number of parameters as \texttt{MACE-MPA-0}~\cite{batatia2023foundation}). The training procedure is done in two stages (similar to how the foundation model is trained from scratch). During the first stage, the forces' weight in the loss computation is way higher than the other predicted targets, then during the second stage, we match the energy weight to that of the forces weight.
    \item We fine-tune that same model separately on the split.
\end{enumerate}
Evaluation results on the test set are reported in Table~\ref{tab:model_performance}. LeMat-Traj helps facilitate effective transfer learning across functionals, especially when data or computational resources are limited, and can help in the development and research of general cross-atomic data source learning methods like \citet{shoghi2023molecules, huang2025cross0functional}. Results show that using a model pre-trained on one functional helps transferring to another functional more easily and in fewer steps.

\begin{table}[htbp]
\centering
\caption{
Performance of pre-trained MACE and ORB Models on Different DFT Functionals split when fine-tuning on a functional split (referred to with \texttt{<split>}) and after fine-tuning (\texttt{-<split>-ft}).
Energy MAE is reported in meV/atom, Force MAE in meV/\AA, Stress MAE in meV/\AA$^3$, and Cosine Similarity is averaged over the forces vectors. All measures are across the test split described in Appendix~\ref{app:experiments}.
}
\label{tab:model_performance}
\resizebox{\textwidth}{!}{%
\begin{tabular}{l
S[table-format=2.2] S[table-format=3.0] S[table-format=1.2] S[table-format=1.3] % PBESol columns
S[table-format=4.0] S[table-format=4.0] S[table-format=-1.3]} % r2SCAN columns
\toprule
\multirow{2}{*}{Model}
& \multicolumn{4}{c}{PBESol}
& \multicolumn{3}{c}{r2SCAN} \\
\cmidrule(lr){2-5} \cmidrule(lr){6-8}
& {Energy MAE} & {Force MAE} & {Stress MAE} & {Cosine Sim.}
& {Energy MAE} & {Force MAE} & {Cosine Sim.} \\
\midrule
\texttt{MACE-MPA-0}        & 370.9  & 101 & 14.7  & 0.13  & 9204.9 & 111 &  0.15 \\
\texttt{MACE-PBESol}       & 51.2  & 33  & 2.1  & 0.04  & {/}  & {/}  & {/}    \\
\texttt{MACE-MPA-0-PBESol-ft} & 18.0 & 27  & 1.6 & 0.19  & {/}  & {/}  & {/}    \\
\texttt{MACE-r2SCAN}       & {/}   & {/} & {/}  & {/}   & 141.7  & 36  &  0.09 \\
\texttt{MACE-MPA-0-r2SCAN-ft} & {/}   & {/} & {/}  & {/}   & 96.3 & 28  &  0.22 \\
\bottomrule
\end{tabular}%
}
\end{table}

\subsection{Limitations and Future Work}

While LeMat-Traj and LeMaterial-Fetcher mark substantial advancements, several areas offer opportunities for improvement.
The current dataset primarily consists of DFT geometry optimization trajectories, and does not include molecular dynamics (MD) trajectories, which could enhance modeling of dynamic properties. Additionally, although the dataset is chemically diverse, the PBE split is largely drawn from the Alexandria database, potentially introducing some data source bias. Future work should aim to incorporate MD trajectories and correctly identify them to diversify data origins, while ensuring compatibility and avoid incorporating noisy data points.
This initial release primarily focuses on dataset construction and characterization; comprehensive benchmarking of MLIPs trained on LeMat-Traj is planned to fully demonstrate its utility (preliminary results in Appendix~\ref{app:experiments}).
Finally, the pipeline in LeMaterial-Fetcher is designed to gather detailed DFT calculation parameters if available from the source (e.g., k-point meshes, pseudopotentials). While not fully exploited in the current version of LeMat-Traj for all entries, this capability can help introduce future MLIP architectures that explicitly embed these parameters as inputs, leading to more versatile multi-fidelity models, enabling LeMat-Traj to continually evolve as a richer resource for the community.
Aggregating data from sources using different underlying DFT parameters (e.g., k-point grids, pseudopotentials) without explicit harmonization risks introducing noise. While we ensure pseudopotential compatibility for included elements following the method in~\citet{siron2025lematbulk}, a deeper quantitative analysis of these potential cross-database biases is an important area for future investigation. We note that LeMaterial-Fetcher's provenance tracking is a first step, enabling researchers to isolate and study these effects.

\section{Conclusion}

In this work, we introduced LeMat-Traj, a scalable, high-quality and unified dataset comprising over 120 million atomic configurations from DFT relaxation trajectories, and LeMaterial-Fetcher, the open-source library enabling its creation and continued evolution. By harmonizing data from prominent repositories across multiple DFT functionals, LeMat-Traj lowers the barrier to training robust, transferable, and accurate MLIPs. Our analysis demonstrates its comprehensive sampling of the potential energy surface along relaxation pathways, capturing both high-energy structures and near-equilibrium states, making it a valuable resource for researchers to develop next-generation interatomic potentials, explore multi-fidelity learning, and advance self-supervised learning techniques in materials science.

While LeMat-Traj currently focuses on geometric optimization trajectories, the modularity of LeMaterial-Fetcher enables future expansions. With the incorporation of compatible molecular dynamics simulations, diversifying data sources further, and implementing dataset-level sampling strategies for more coherent fine-tuning datasets. Integrating LeMaterial-Fetcher with automated active learning and DFT calculation workflows can enable the continuous enrichment of LeMat-Traj with high-fidelity data. We believe LeMat-Traj and LeMaterial-Fetcher represent a step towards democratizing access to high-quality, curated training data, fostering community collaboration, and ultimately accelerating the pace of data-driven materials discovery

\section*{Acknowledgments}
This project was provided with computer and storage resources by GENCI at
CINES and IDRIS under the allocations 2024-AD011015800, 2025-A0181016212 and 2025-AD011016353 on the supercomputers
Jean Zay and Adastra's A100/MI300 partition.

% \begin{ack}
% Thank LeMaterial contributors, dataset sources, JZ for compute, whatelse?
% \end{ack}

\bibliographystyle{plainnat}
\bibliography{references}

%%%%%%%%%%%%%%%%%%%%%%%%%%%%%%%%%%%%%%%%%%%%%%%%%%%%%%%%%%%%

\appendix

\section{Data Availability and Licensing}
LeMat-Traj is publicly available at \url{https://huggingface.co/datasets/LeMaterial/LeMat-Traj} and is distributed under the Creative Commons Attribution 4.0 International (CC-BY 4.0) license. LeMaterial-Fetcher library, developed for the curation of LeMat-Traj, is open-source and available on GitHub at \url{https://github.com/LeMaterial/lematerial-fetcher}. LeMaterial-Fetcher is distributed under the Apache License 2.0.

LeMat-Traj aggregates, filters and standardizes data from the following publicly available repositories:
\begin{itemize}
    \item \textbf{The Materials Project} \citep{jain2013commentary, jain2020materials}
    \item \textbf{Alexandria} \citep{schmidt2021alexandria, schmidt2024improving}
    \item \textbf{The Open Quantum Materials Database (OQMD)} \citep{saal2013materials}
\end{itemize}

All data retrieved from these original sources for inclusion in LeMat-Traj are distributed under licenses compatible with CC-BY 4.0, primarily their own CC-BY 4.0 licenses. Specifically, for data originating from the Materials Project, care was taken to ensure that only structures and calculations designated under the CC-BY 4.0 license were included. We gratefully acknowledge the original creators and maintainers of these foundational datasets for making their valuable work publicly accessible.

\section{Distribution Analysis}

% We illustrate the data provenance of atomic configurations in Figure~\ref{fig:dataset_source}.

% \begin{figure}
%     \centering
%     \includegraphics[width=0.6\linewidth]{figures/frames_distribution.pdf}
%     \caption{Visual Representation of Table~\ref{tab:functional_data_summary}.}
%     \label{fig:dataset_source}
% \end{figure}

\begin{figure}[h]
  \centering
    \includegraphics[width=1\linewidth]{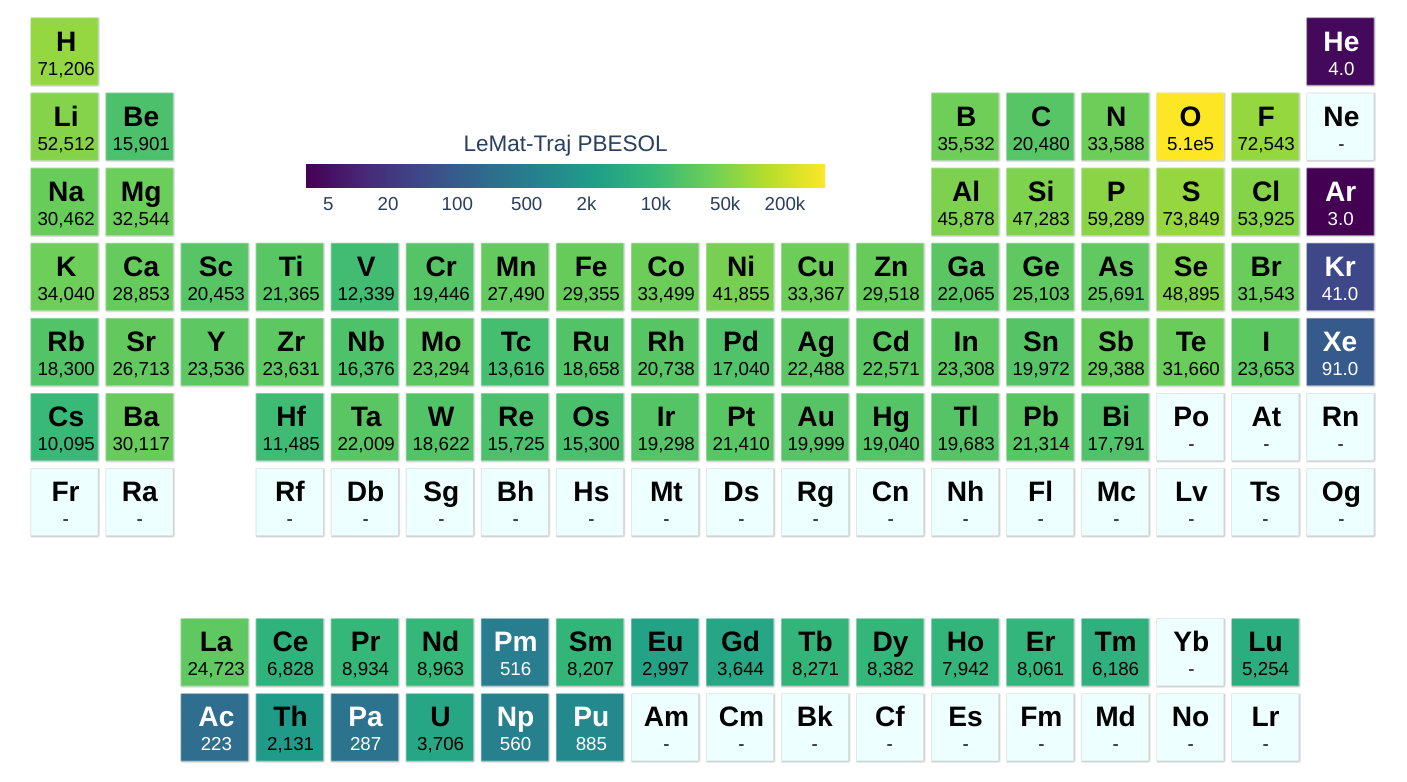}
  \caption{Chemical distribution in number of trajectories for the PBESol split.}
  \label{fig:chemical_formula_distribution_pbesol}
\end{figure}

\paragraph{Chemical diversity.}
To highlight the chemical diversity of the dataset, Figure~\ref{fig:chemical_formula_distribution_pbesol} and \ref{fig:chemical_formula_distribution} present periodic table heatmaps of the number of trajectories involving each element for the LeMat-Traj dataset, separately for the PBE and PBESol splits. The distribution spans nearly the entire periodic table, with particularly high representation of elements such as transition metals (e.g., Fe, Ni, Co), light elements (e.g., H, C, O, N), and main group elements (e.g., Si, Al, S). Besides oxides dominating and actinides being under-represented, the distribution is well-balanced. This ensures that the dataset is suitable for training universal machine-learned interatomic potentials that generalize across diverse chemistries and bonding environments.

\begin{figure}[htbp]
    \centering
    \includegraphics[width=0.7\linewidth]{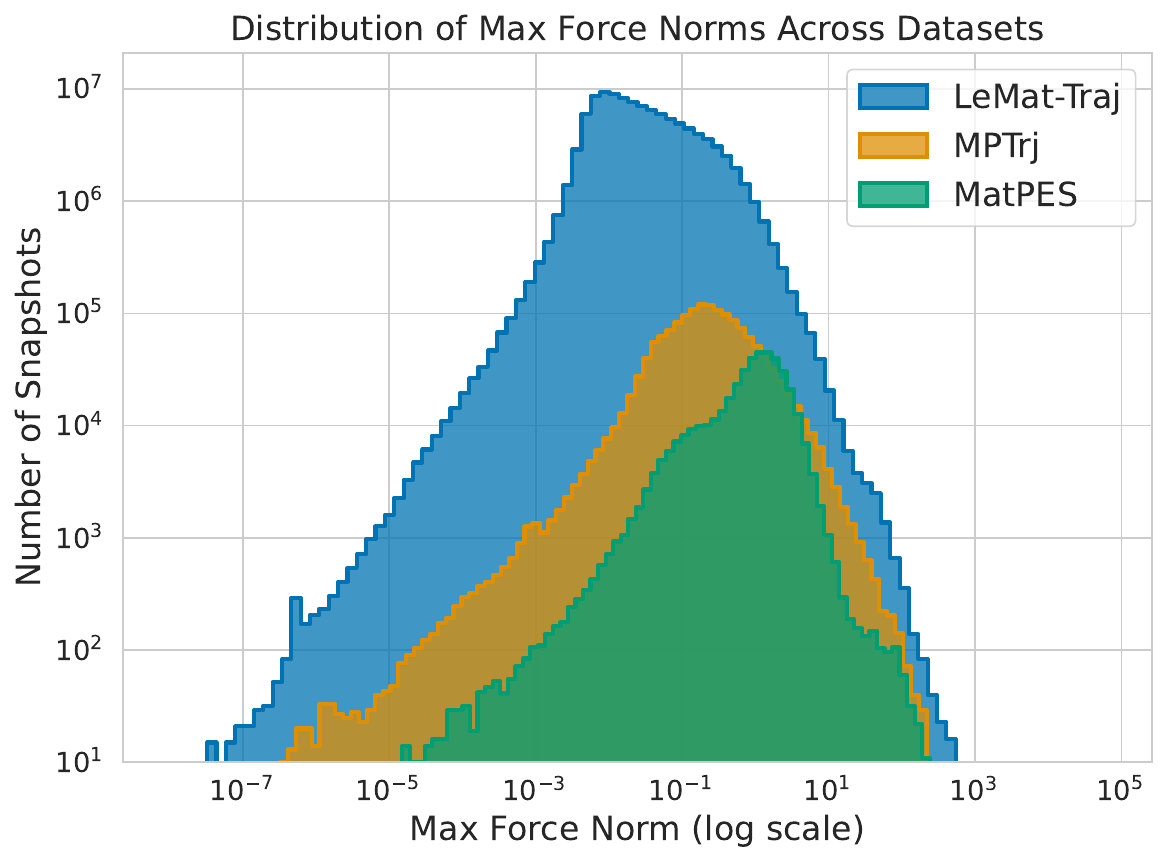}
    \caption{Coverage in log-log scale of the maximum norm of the force vector on every atomic configurations in LeMat-Traj (PBE split), MPtrj and MatPES.}
    \label{fig:max_force_norm}
\end{figure}

\paragraph{Max Force.}
Figure~\ref{fig:max_force_norm} displays the distribution of maximum atomic force norms, revealing LeMat-Traj's (PBE split) extensive coverage. It contains substantially more configurations spanning a wider range of force magnitudes (from approximately $10^{-7}$ to $10^3$ eV/Å) compared to MPTrj and MatPES, indicating comprehensive sampling from near-equilibrium to high-force states.

\begin{figure}[htbp]
    \centering
    \includegraphics[width=1\linewidth]{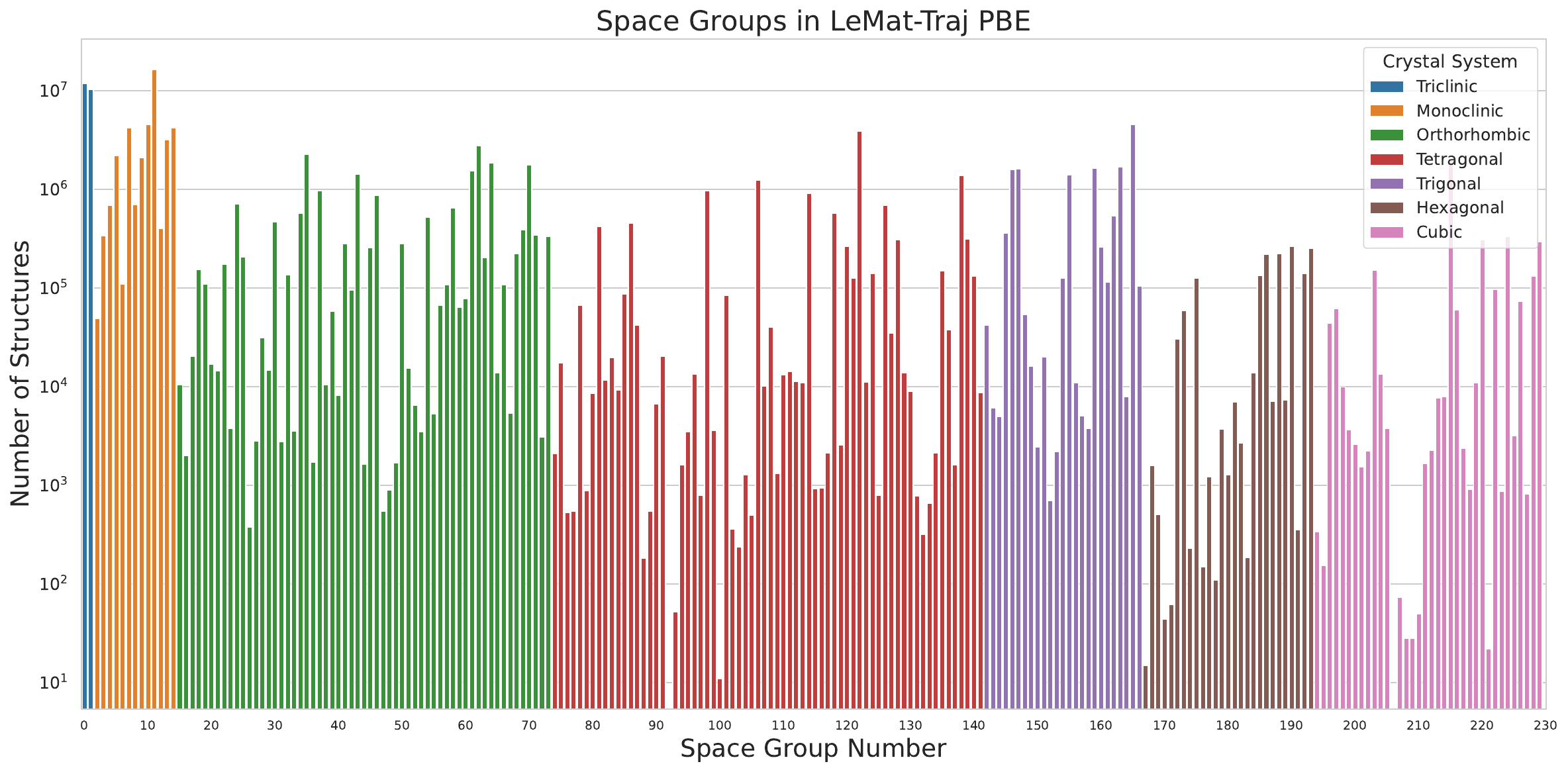}
    \caption{
    Distribution of space groups in LeMat-Traj (PBE subset), categorized by crystal system. The figure illustrates the number of structures for each space group on a logarithmic scale, highlighting the dataset’s broad coverage of crystallographic symmetries. All seven crystal systems are represented, spanning over 200 distinct space groups.
    }
    \label{fig:space_groups}
\end{figure}

\paragraph{Space Group diversity.} To assess the structural diversity of the dataset, we analyzed the distribution of crystallographic space groups for the LeMat-Traj PBE subset. 
The space groups of the 120M structures were computed during the dataset creation using \texttt{moyo} a faster alternative to \texttt{Spglib}~\citep{togo2024textttspglibsoftwarelibrarycrystal} in \texttt{LeMaterial-Fetcher}. The strict default parameters for space group identification (\texttt{symprec} $10^{-4}$) were used in the dataset, allowing for a unified space group description across all the structures.
As shown in Figure~\ref{fig:space_groups}, the dataset spans the full range of crystal systems, including triclinic, monoclinic, orthorhombic, tetragonal, trigonal, hexagonal, and cubic groups. More than 200 unique space groups are represented, with a significant number of entries in low-symmetry systems (e.g., triclinic and monoclinic), which can be explained by the strict tolerance. This symmetry diversity is essential for training machine learning interatomic potentials (MLIPs) that generalize across materials with varying spatial constraints and bonding environments. It is also worth noting that 98\% of the trajectories are assigned the same space group label at the first step of the relaxation and the last one showing the symmetry conservation during the geometric optimization calculations.

\begin{figure}[htbp]
    \centering
    \includegraphics[width=1\linewidth]{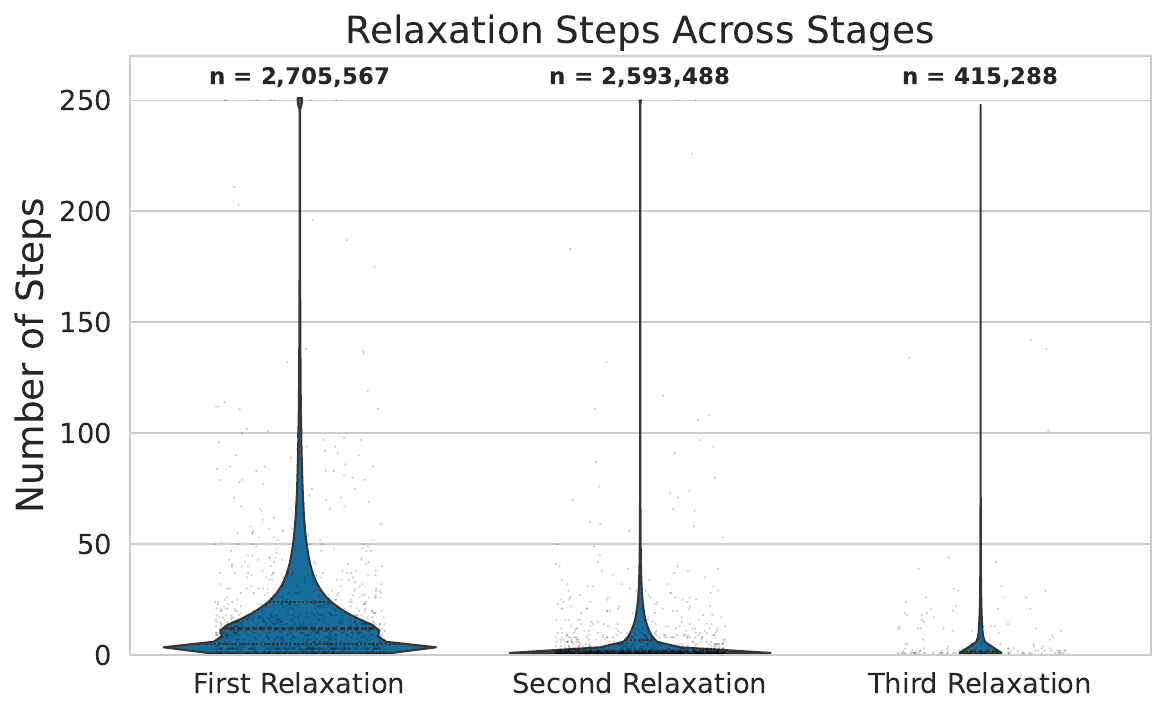}
    \caption{
     Number of geometry optimization steps across the first, second, and third relaxation stages in LeMat-Traj. The density of number of steps for each stage, with the total number of trajectories (n) labeled above are represented. While the first relaxation often involves more extensive structural changes, subsequent stages typically require fewer steps, indicating convergence toward optimized geometries.
    }
    \label{fig:relaxation_numbers}
\end{figure}

\paragraph{Relaxation Steps.}
Figure~\ref{fig:relaxation_numbers} illustrates the distribution of the number of geometry optimization steps performed across the first, second, and third relaxation stages within LeMat-Traj as described in section~\ref{sec:relaxation_number}. The plots reveal that the first relaxation generally involves a broader and more varied distribution of steps, often exceeding 50 or even 100 steps for more complex or strained initial structures. In contrast, the second and third relaxations show sharply peaked distributions concentrated at lower step counts, reflecting incremental refinements of already partially relaxed geometries. This progression highlights the effectiveness of multi-stage relaxation strategies in achieving convergence, while also emphasizing that the dataset captures a wide range of relaxation behaviors—from flat minima to deep, multi-step optimization paths.

\section{Experiments on LeMat-Traj}
\label{app:experiments}

\subsection{Subsets of LeMat-Traj.} 
\label{app:subset}
In this section, we provide additional details on the way the subsets of LeMat-Traj were created and splitted for the small experiments. Due to the dataset's size, we focus on measuring performances on a few selected subsets of the dataset. The splits are available at \url{https://huggingface.co/datasets/LeMaterial/LeMat-Traj-subset} and can be used on more limited computational resources. Each entry represents an atomic configuration within a trajectory. To avoid data leakage, subsampling and splitting are performed at the trajectory level, ensuring all configurations from a given trajectory appear exclusively in either the training or test set. Splits are stratified based on the one-hot encoding of chemical elements present in the trajectory. This ensures no atomic species in the test set are unseen during training—essential for model generalizability.
To ensure balance between the different sources for all subsets, we keep the same 10\% MP, 10\% OQMD and 80\% Alexandria balance across all splits and all functionals, as long as the data source provides data for the functional. For SCAN and r2SCAN where the only provenance source is Materials Project, we keep all the data from the original dataset in these subset because they are small enough for these experiments and split the train and test split with a stratified 80-20\% separation of the trajectories.

\subsection{Cross-Dataset Generalization}
The benchmarks in Section \ref{sec:benchmarks} highlight that combining high-force data (OMat24) with near-equilibrium data (LeMat-Traj) yields the best performance. To further explore this, we conducted a cross-dataset evaluation, testing models trained on one dataset against the test sets of others. As shown in Tables \ref{tab:matpes_test}, \ref{tab:omat24_test}, and \ref{tab:mptrj_test}, models consistently perform best on their in-distribution test data. For example, the model trained on OMat24 achieves the lowest errors on the OMat24 test set, but performs poorly on the LeMat-Traj test set (Table \ref{tab:finetuning_benchmark}), and vice-versa. This reinforces our central argument: different data generation strategies (MD/active learning vs. geometry optimization) capture distinct but complementary regions of the potential energy surface. A single data source is often insufficient for creating a truly general-purpose potential. Our results demonstrate that LeMat-Traj is a crucial resource for specializing models in the low-force regime essential for accurate relaxations, complementing existing high-force datasets.

\begin{table}[h]
\centering
\caption{Evaluation on the MatPES PBE 10K held-out test set.}
\label{tab:matpes_test}
\begin{tabular}{lccc}
\toprule
\textbf{Training Dataset} & \textbf{Energy MAE (meV) ↓} & \textbf{Force MAE (meV/Å) ↓} & \textbf{Force Cos ↑} \\
\midrule
OMat24 & 193.8 & \textbf{123.5} & 0.77 \\
MPtrj & 250.2 & 187.5 & 0.70 \\
MatPES PBE & \textbf{56.6} & 127.1 & \textbf{0.78} \\
LeMat-Traj only & 245.8 & 217.9 & 0.68 \\
OMat24 + ft LeMat-Traj & 249.1 & 203.9 & 0.75 \\
\bottomrule
\end{tabular}
\end{table}

\begin{table}[h]
\centering
\caption{Evaluation on the OMat24 Validation 10K test set.}
\label{tab:omat24_test}
\begin{tabular}{lccc}
\toprule
\textbf{Training Dataset} & \textbf{Energy MAE (meV) ↓} & \textbf{Force MAE (meV/Å) ↓} & \textbf{Force Cos ↑} \\
\midrule
OMat24 & \textbf{17.9} & \textbf{103.4} & \textbf{0.99} \\
MPtrj & 156.4 & 404.5 & 0.94 \\
MatPES PBE & 312.3 & 358.8 & 0.96 \\
LeMat-Traj only & 153.6 & 598.3 & 0.95 \\
OMat24 + ft LeMat-Traj & 218.5 & 395.8 & 0.97 \\
\bottomrule
\end{tabular}
\end{table}
\begin{table}[h]
\centering
\caption{Evaluation on the MPtrj 10k held-out test set.}
\label{tab:mptrj_test}
\begin{tabular}{lccc}
\toprule
\textbf{Training Dataset} & \textbf{Energy MAE (meV) ↓} & \textbf{Force MAE (meV/Å) ↓} & \textbf{Force Cos ↑} \\
\midrule
OMat24 & 58.7 & 68.7 & \textbf{0.54} \\
MatPES PBE & 237.6 & 114.6 & 0.36 \\
LeMat-Traj only & \textbf{20.2} & \textbf{63.3} & 0.52 \\
OMat24 + ft LeMat-Traj & 37.3 & 73.4 & 0.52 \\
\bottomrule
\end{tabular}
\end{table}

\subsection{Model Training.}
We report in Table~\ref{tab:finetuning_hyperparams} the hyperparameters used for training MACE. Experiments were all conducted on a single A100-40GB GPU.

\begin{table}[h]
\centering
\caption{Hyperparameters used to train MACE on the subsets of LeMat-Traj.}
\label{tab:finetuning_hyperparams}
\begin{tabular}{lccc}
\toprule
\textbf{Hyperparameter} & \textbf{Training Stage 1} & \textbf{Training Stage 2} & \textbf{Fine-tuning} \\
\midrule
Learning Rate & 8e-4 & 8e-4 & 8e-4 \\
Scheduler & Constant & Constant & Constant \\
Batch Size & 128 & 128 & 128 \\
Energy Weight & 1 & 100 & 1 \\
Force Weight & 10 & 100 & 100 \\
Stress Weight & 1 & 1 & 1 \\
\bottomrule
\end{tabular}
\end{table}

\section{LeMaterial-Fetcher}
\label{app:lemat_fetcher}
As described in section~\ref{sec:relaxation_number}, the pipeline to download and process the datasets is made to be both extremely customizable but also highly parallel and scalable. By default, LeMaterial-Fetcher uses \texttt{PostgreSQL} as a backend to dump the raw downloaded datasets but also to process the transformed structures before pushing them to HuggingFace. Other backends are supported and easy to integrate in the library, with for example \texttt{MySQL} being used for OQMD (the source dataset from their website is a full database with scattered tables).
One of the main challenges with writing this pipeline was allowing for full parallelization to decrease the time from download to pushing the unified dataset. 
Indeed, having multiple connections opened for both fetching data from a table and pushing them to the other one with database cursors is prone to high memory usage and leakage. Naive implementations of parallelism do not allow to fully take advantage of high compute machines. To that end, we designed the library to be very memory-efficient. For LeMat-Traj, it was possible to take advantage of 128 cores with 256GB without any issue.
The entire pipeline to create LeMat-Traj took around 16 hours to create the 120M rows and upload them on HuggingFace running with 12 workers on an AMD Ryzen 5600G. This time gets significantly reduced when running on larger machine on which we are able to max-out the usage. \\
For the dataset curation process, we follow the same procedure as \citep{siron2025lematbulk} with the exception that we pick Ytterbium (Yb) containing samples from Materials Project rather than Alexandria because of the non-compatibility between their pseudo-potentials.

\paragraph{Materials Project.}
For the Materials Project data transformation process, we look through every single task available (around 1.5M at the latest release during the first LeMat-Traj version), and then only keep the non-deprecated tasks. To ensure accurate sampling of the PES, we pick all the trajectories for a given material as long as they pass the data filtering described in \ref{sec:data_filtering}.

\paragraph{Alexandria.}
All samples from Alexandria were used except for the ones containing Yb.

\paragraph{OQMD.}
OQMD trajectories are obtained by going through all the entries of the OQMD database, gathering their associated calculations from \textit{relaxation}, \textit{coarse relaxation} and \textit{fine relaxation} for every relaxation stage. The input structures and output structures are then processed, provided they contain the targets expected in the right format.

\section{Potential Energy Surfaces}
\label{app:pes}
\begin{figure*}[htbp]
  \centering
  \begin{subfigure}{0.475\linewidth}
    \includegraphics[width=\linewidth]{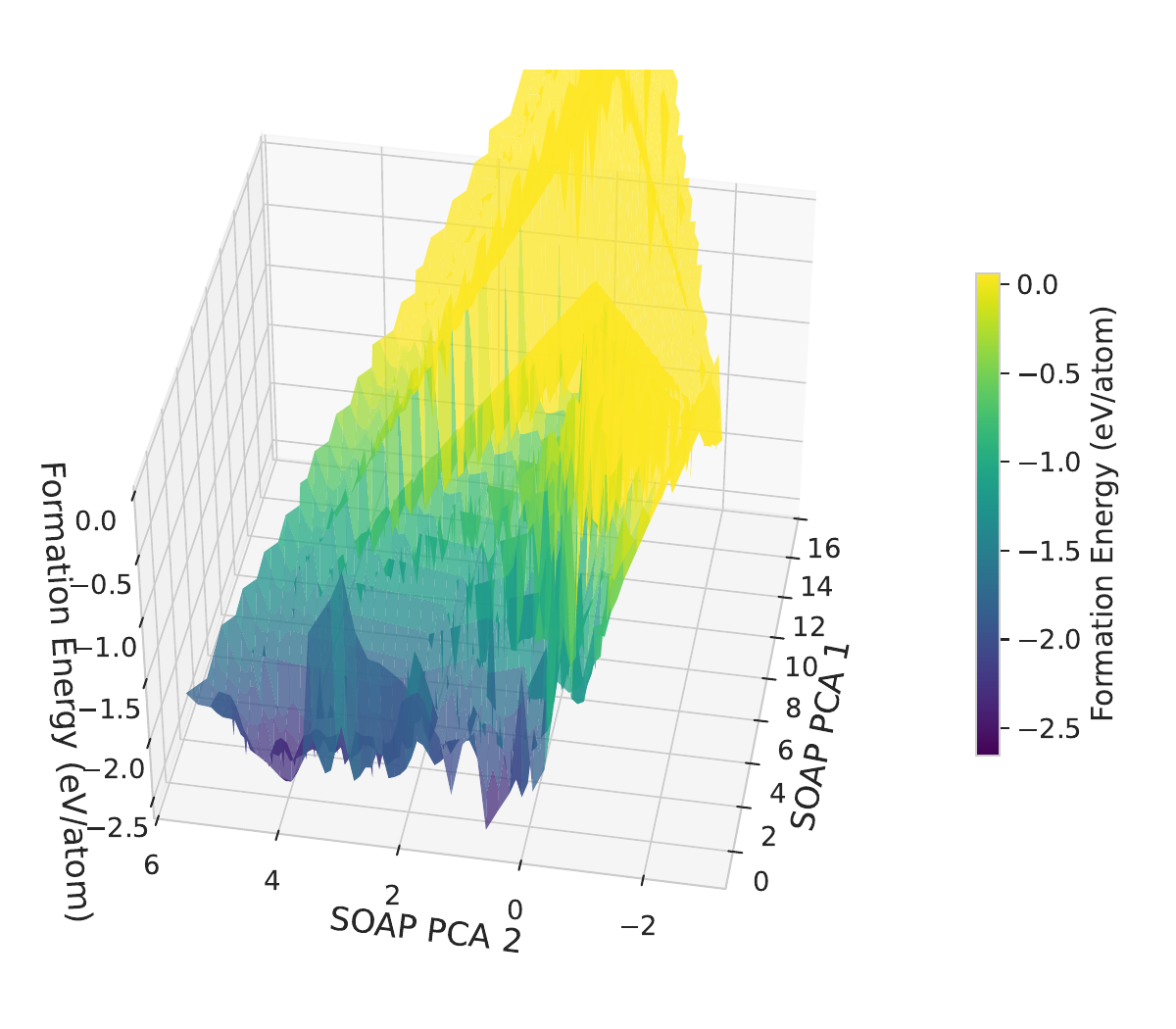}
    \caption{LeMat-Traj PBE}
    \label{fig:lemat_pes_3d_na_cl_o}
  \end{subfigure}
  \hfill
  \begin{subfigure}{0.475\linewidth}
    \includegraphics[width=\linewidth]{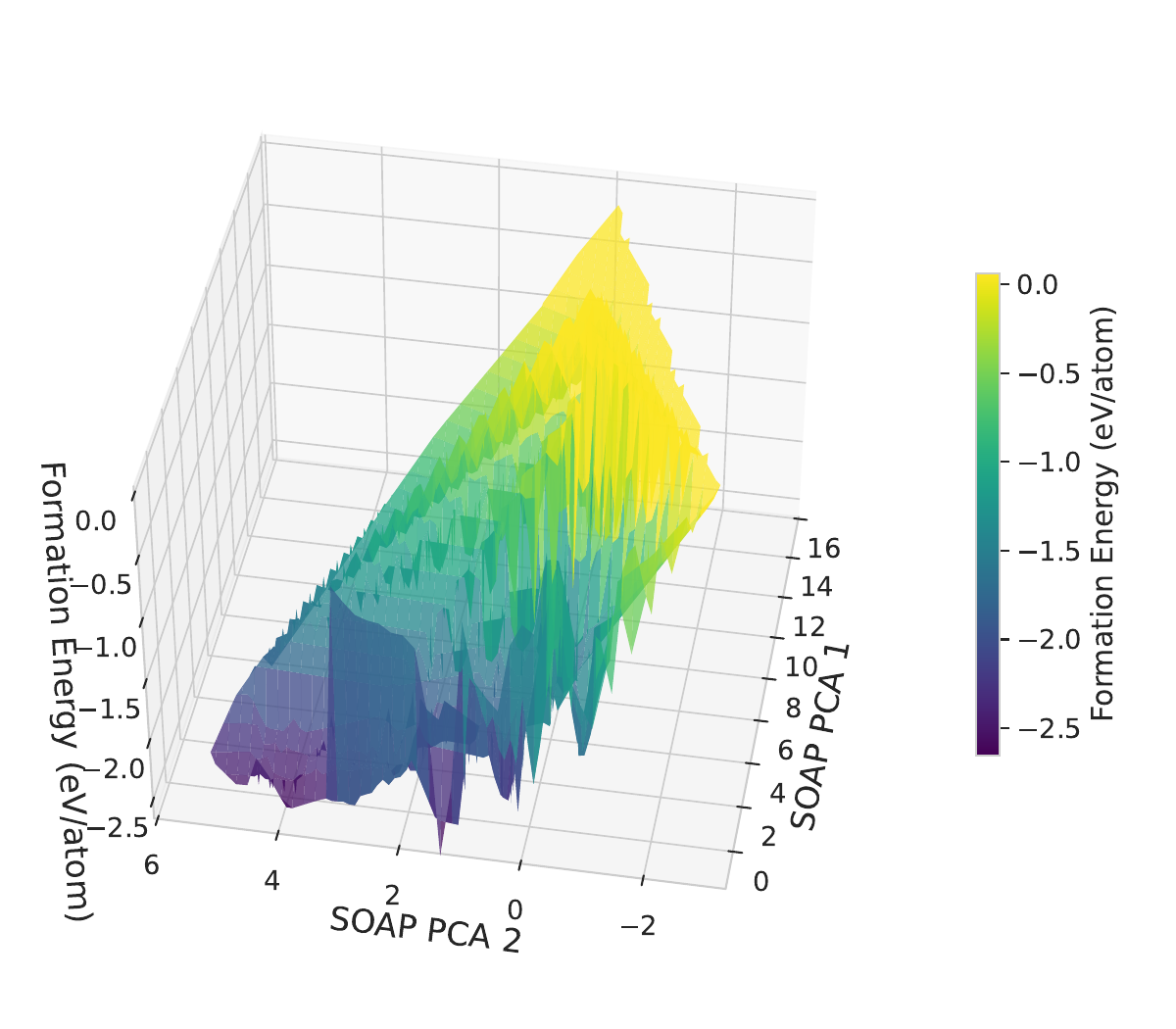}
    \caption{MPtrj}
    \label{fig:mptrj_pes_3d_na_cl_o}
  \end{subfigure}
  \caption{Projected Potential Energy Surfaces (PES) for the ionic Na-Cl-O systems for LeMat-Traj and the MPtrj datasets, similar to Figure~\ref{fig:pes_comparison} in 3D projection.}
  \label{fig:mptrj_pes_na_cl}
\end{figure*}

In order to get a more visual understanding of the differences between existing datasets, we make an attempt at a framework to plot the Potential Energy Surface in principal components. To allow for easier interpretability we limit the analysis to specific coherent subsets of chemical elements (metallic or ionic). For every dataset, all the atomic configurations whose chemical formula is a subset of the chosen elements are gathered. Then SOAP descriptors are computed for all these configurations with the same hyperparameters (\texttt{r\_cut} = 5.0, \texttt{n\_max} = 8 and \texttt{l\_max} = 6, with \texttt{outer} averaging to get a vector for every structure). All of these SOAP vectors are used to fit a PCA and the formation energy per atom (eV/atom) is computed. Because the sampling of atomic configurations is scattered across the PCA space and not continuous, we use a linear interpolation of the convex hull to get this visual description.
Figure~\ref{fig:mptrj_pes_na_cl} illustrates the PES of a different chemical subset, highlighting the close similarity between LeMat-Traj and MPtrj. Indeed, since MPtrj is contained in LeMat-Traj, the PES of the latter describes local minima and transition pathways with a higher resolution. Additionally, when only limiting the sampling to two elements systems with Fe-Cu, we notice the advantages of having a larger structural configuration sampling to better describe the entire PES. Although having a smaller dataset may result in a smoother landscape that might help models converge faster and more easily, it is not enough to completely capture the large number of local energy minima that exist in the complex DFT force field.

\begin{figure*}[hbtp]
  \centering
  \begin{subfigure}{0.475\linewidth}
    \includegraphics[width=\linewidth]{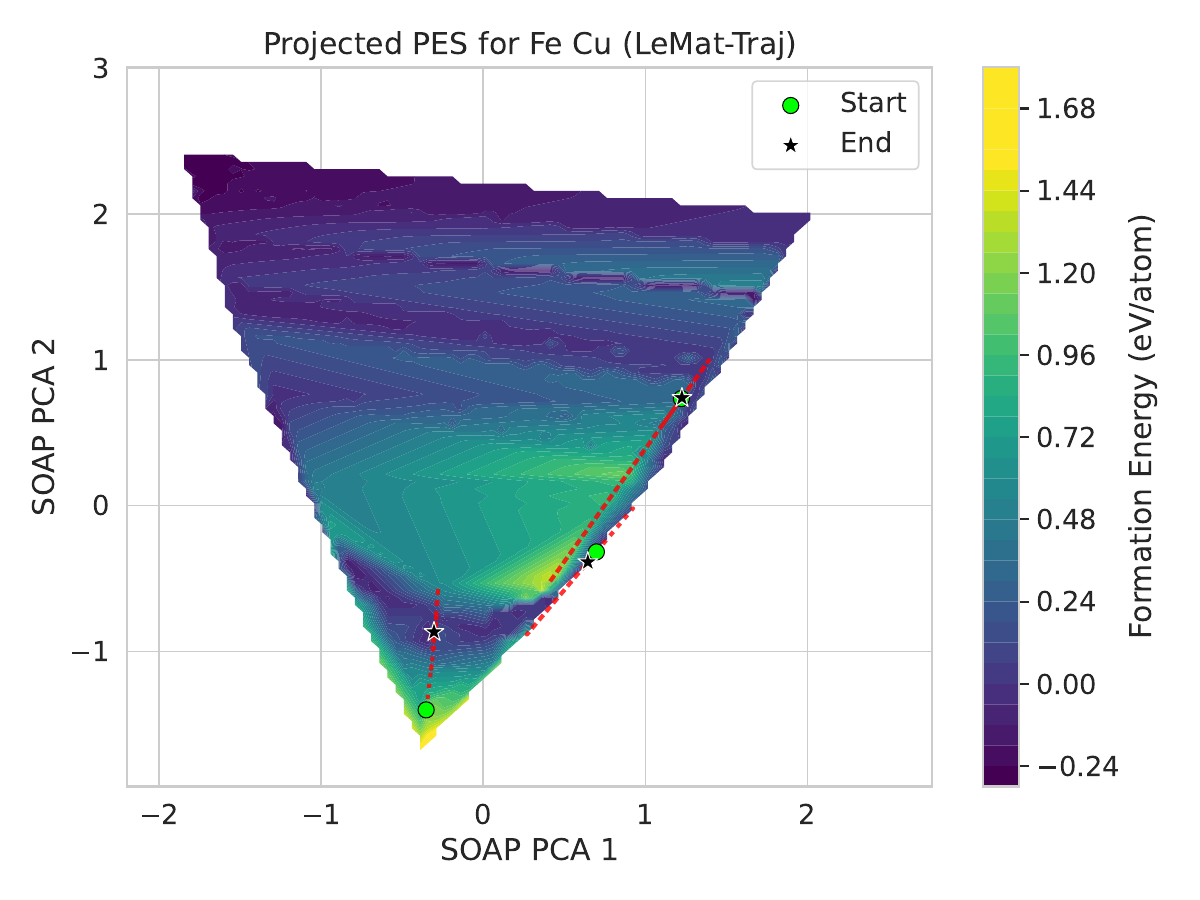}
    \caption{LeMat-Traj PBE}
    \label{fig:lemat_pes_2d_fe_cu}
  \end{subfigure}
  \hfill
  \begin{subfigure}{0.475\linewidth}
    \includegraphics[width=\linewidth]{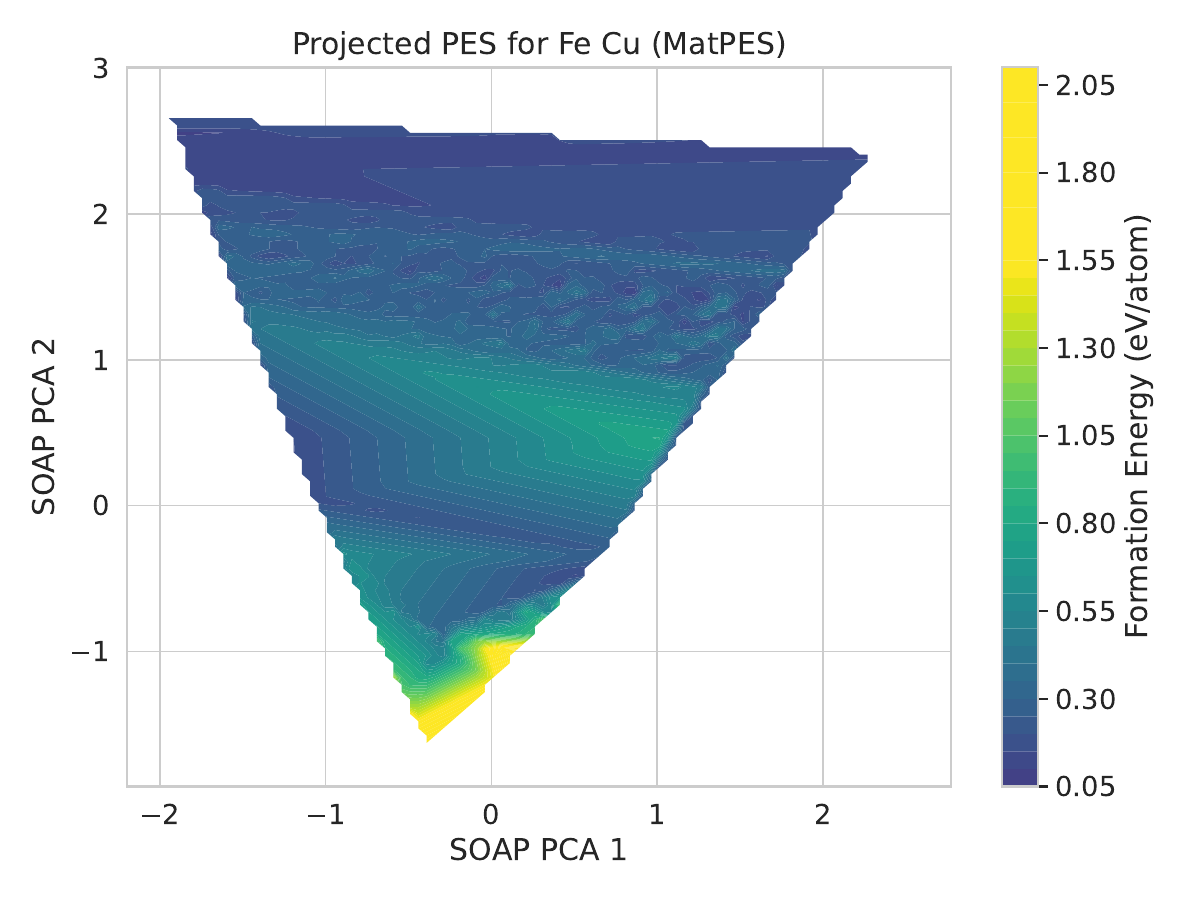}
    \caption{MatPES}
    \label{fig:matpes_pes_2d_fe_cu}
  \end{subfigure}
  \caption{Projected Potential Energy Surfaces (PES) for the subset Fe-Cu systems for LeMat-Traj and the MPtrj datasets, similar to Figure~\ref{fig:pes_comparison}.}
  \label{fig:pes_fe_cu_comparison}
\end{figure*}

\end{document}